\documentclass[letterpaper, 10 pt, conference]{ieeeconf}  

\IEEEoverridecommandlockouts                              

\usepackage{graphicx} 
\usepackage{amsmath} 
\usepackage{mathtools}
\usepackage{amsfonts}
\usepackage{cases}
\usepackage{nameref}
\usepackage{varioref}
\usepackage{hyperref}
\usepackage{cleveref}
\usepackage[overload]{empheq}
\newcommand{\com}{\boldsymbol{x}}
\newcommand{\dcm}{\boldsymbol{\xi}}
\newcommand{\vrp}{\boldsymbol{r}_{vrp}}
\newcommand{\vrpo}{\boldsymbol{r}_{vrp,0}}
\newcommand{\vrpT}{\boldsymbol{r}_{vrp,T}}

\newcommand{\vrpt}{\boldsymbol{r}_{vrp,t}}

\newcommand{\vproj}{\Tilde{\boldsymbol v}_{t}}
\newcommand{\dcmproj}{\Tilde{\dcm}_{t}}
\newcommand{\g}{\boldsymbol{g}}
\newcommand{\cont}{\boldsymbol{p}}
\newcommand{\force}{\boldsymbol{f}}
\newcommand{\angmom}{\boldsymbol{L}}
\newcommand{\cop}{\boldsymbol r_{cop}}
\newcommand{\nextStep}{\boldsymbol{u}_T}
\newcommand{\curStep}{\boldsymbol{u}_0} 
\newcommand{\delu}{\Delta \boldsymbol{u}}

\newtheorem{remark}{\textbf{Remark}}

\title{\LARGE \bf
A unified framework for walking and running of bipedal robots
}

\author{Mahrokh Ghoddousi Boroujeni$^{1*}$, Elham Daneshmand$^{2*}$, Ludovic Righetti$^{2,3}$, Majid Khadiv$^{2}$
\thanks{}
\thanks{This work was supported by the Max-Planck Institute for Intelligent Systems' Grassroots program (M10338 and M10343), New York University, the European Union’s Horizon 2020 research and innovation program (grant agreement 780684) and the National Science Foundation (CMMI-1825993).}
\thanks{$^{*}$ Authors equally contributed to this work.}
\thanks{$^{1}$ Institute of Mechanical Engineering, École Polytechnique Fédérale de Lausanne (EPFL), Switzerland.}
\thanks{$^{2}$Max Planck Institute for Intelligent Systems, Tübingen, Germany.}%
\thanks{$^{3}$ Tandon School of Engineering, New York University, Brooklyn, USA.}
}

\begin{document}

\maketitle
\thispagestyle{empty}
\pagestyle{empty}

\begin{abstract}
In this paper, we propose a novel framework capable of generating various walking and running gaits for bipedal robots. The main goal is to relax the fixed center of mass (CoM) height assumption of the linear inverted pendulum model (LIPM) and generate a wider range of walking and running motions, without a considerable increase in complexity. To do so, we use the concept of \emph{virtual constraints} in the centroidal space which enables generating motions beyond walking while keeping the complexity at a minimum. 
By a proper choice of these virtual constraints, we show that we can generate different types of walking and running motions.
More importantly, enforcing the virtual constraints through feedback renders the dynamics linear and enables us to design a feedback control mechanism which adapts the next step location and timing in face of disturbances, through a simple quadratic program (QP). 
To show the effectiveness of this framework, we showcase different walking and running simulations of the biped robot Bolt in the presence of both environmental uncertainties and external disturbances.
\end{abstract}
\section{INTRODUCTION}
Legged robots can perform a wide range of complex maneuvers through synchronous joint motions that satisfy very limiting contact interaction constraints. However, the underlying dynamics of these robots are highly nonlinear and hybrid that renders the problem of generating and controlling these different motions highly complicated. Traditionally, common practice has been to have various template models to control different motions, e.g., the linear inverted pendulum model (LIPM) \cite{kajita20013d} for walking and the spring-loaded inverted pendulum (SLIP) \cite{blickhan1989spring} for running. While using these models to generate plans and controllers for locomotion has led to promising results \cite{takenaka2009real,hubicki2018walking,daneshmand2021variable}, little effort has been dedicated to their unification through a more formal and general framework \cite{chen2020optimal}.

More recently, the community has tried to use more general models that enable generating a more comprehensive range of motions through the use of the centroidal momentum dynamics \cite{herzog2016structured,dai2016planning,carpentier2018multicontact}. Centroidal momentum dynamics capture the relationship between external forces and centroidal states, \emph{exactly} \cite{wieber2006holonomy}; hence it can be seen as a general model for locomotion problems. However, since the dynamics are nonlinear, this gives rise to non-convex optimization problems that need to solve several convex sub-problems for converging to a local minimum. While very efficient solvers for this problem have been developed \cite{carpentier2018multicontact,ponton2021efficient,sleiman2021unified,shah2021rapid}, they are still at least one order of magnitude slower than template-based approaches \cite{daneshmand2021variable}, and in general, they can provide little guarantees in terms of constraint satisfaction and convergence to a \emph{good} local minimum.

Another model-based approach for controlling legged robots relies on the (hybrid) zero dynamics (HZD) concept \cite{westervelt2018feedback}. In this approach, all the computations required for generating a cyclic gait based on whole-body dynamics are carried out offline through a highly non-convex optimization problem that usually minimizes energy consumption \cite{reher2016realizing}. The main idea in this framework is to use \emph{virtual constraints} for the desired joints motions (or some important points on the robot such as ankle, base, etc.,) parameterized by Bezier polynomials and to optimize the boundary conditions of the Bezier curve such that the resulting motion is periodic. Feedback linearization is then used to ensure that the resulting lower dimensional dynamics (zero dynamics) are attractive and invariant under the continuous time dynamics of the system. Since all the computations for generating periodic gaits are carried out offline, a library of gaits is developed for different walking/running speeds. Then, they are interpolated such that the resulting gaits also satisfy constraints and guarantee invariance, for instance, using control barrier functions \cite{nguyen2020dynamic} or control Lyapunov functions \cite{reher2021control} through a quadratic programming (QP)-based inverse dynamics framework. While this approach gives impressive formal guarantees and has shown experimental success \cite{sreenath2013embedding,reher2016realizing}, its effectiveness is highly dependant on the quality of the solution provided by the constrained nonlinear optimization problem. Again, little can be asserted in advance about general constraint satisfaction and convergence of this non-convex optimization problem.

In this paper, we aim to formalize the use of template models through the notion of \emph{virtual constraints in the centroidal space}. Note that virtual constraints are different from physical constraints in that they are enforced through the use of feedback \cite{ameshybrid}. To this end, we propose an approach to design natural and gait-dependant virtual constraints that are parameterized using intuitive parameters and render the problem of step location and timing adjustment a simple quadratic program (QP) for both bipedal walking and running. This controller guarantees weak forward invariance (viability) of the gait by limiting the distance between the divergent component of motion (DCM) and next step location inside the viability kernel.  Within this framework, we propose a unified control approach capable of realizing various bipedal walking and running gaits. In contrast with \cite{nguyen2020dynamic,reher2016realizing,guo2021fast}, we do not need to generate offline a gait library and then interpolate between them online, but we can generate walking and running at different desired velocities on the fly. The main contributions of the paper are as follows:

\begin{itemize}
    \item We formalize the use of template models for motion planning for bipedal robots through the concept of virtual constraints in the centroidal space,
    \item We present a fast feedback controller that adapts both step locations and duration in response to external disturbances for both walking and running motions,
    \item We show that we can generate and transition between different walking and running motions on the biped robot Bolt \cite{daneshmand2021variable} in simulation, in the presence of both external disturbances and irregularities of the ground height.
\end{itemize}

\subsection{Definitions and notations}
Each step consists of two phases:
the stance phase with a duration of $T_s > 0$ when one foot is in contact with the ground, and possibly the flight phase when both feet are lifted for $T_f \geq 0$. 
Note that there is no double-support phase.     
Walking does not include a flight phase and the next foot to land is the one raised before.
The stance and swing foot alternate in running and there is a non-zero flight duration. 

\section{Fundamentals}

\subsection{Centroidal momentum dynamics}
The centroidal momentum dynamics capture the relationship between the external forces and the centroidal states, i.e., CoM states and angular momentum \cite{wieber2006holonomy}:
\begin{subequations}
\label{eq:newton-euler}
    \begin{align}
        &\sum_{i=1}^{n} \force_i = m (\ddot \com + \g)  \label{eq:newton},\\
        &\sum_{i=1}^{n} (\cont_i - \com) \times \force_i = \dot{\angmom}, \label{eq:euler}
    \end{align}
\end{subequations}
where $\force_i$ is the external force exerted by the end-effector $i$, $\com = [x, y, z]^T$ is the CoM position, $m$ is the robot mass, and $\g=[0,0,-g]^T$ is the gravity vector.  $\cont_i$ stands for the point of action of force from the end-effector $i$, and $\angmom=[l_x, l_y, l_z]^T$ is the angular momentum around the CoM \cite{orin2013centroidal}.

We focus on walking and running motions in this paper and exclude the multi-contact case from our analysis. Combining \eqref{eq:newton} in $z$ direction with \eqref{eq:euler} in $x,y$ directions, we end up in the following set of equations \cite{wieber2016modeling}:
\begin{subequations}
\label{eq:cop-com-dynamics}
    \begin{align}
        &(z - z_{cop}) \ddot{x} = (\ddot{z} + g) (x - x_{cop})-\frac{\dot{l}_y}{m},\\
        &(z - z_{cop}) \ddot{y} = (\ddot{z} + g) (y - y_{cop})+\frac{\dot{l}_x}{m}.
    \end{align}
\end{subequations}
In the above, $\cop=[x_{cop}, y_{cop}, z_{cop}]^T$ is the center of pressure (CoP) of the foot in contact with the ground surface. In the vertical direction, force balance imposes that
\begin{eqnarray}
    \ddot{z} = f_z/m - g\label{eq:z_force_balance},
\end{eqnarray}
where $f_z$ is the vertical component of the contact forces.

\subsection{Virtual constraints}
This section introduces virtual constraints (that need to be enforced later using feedback control), enabling us to unify motion generation and control for both walking and running of bipedal robots. The considered virtual constraint is to enforce the external forces to point towards the CoM,
\begin{eqnarray}
\label{eq:f_to_com}
    \boldsymbol f = m s (\com - \cop).
\end{eqnarray}
The constant $s$ in \eqref{eq:f_to_com} is positive by the unilaterally of contact forces, and hence, replaced by $s\coloneqq\omega^2$. This virtual constraint gives us two appealing features; first, as the external force is directed from the CoP towards the CoM, the rate of change of angular momentum around the CoM is zero, i.e.,
\begin{equation}
\label{eq:zero-ang-mom}
    \dot l_x = \dot l_y = 0.
\end{equation}
Given \eqref{eq:zero-ang-mom} and the fact that we do not want to have a constant non-zero angular momentum around the CoM, \eqref{eq:cop-com-dynamics} becomes:
\begin{subequations}
\label{eq:cop-com-dynamics-w/o-ang-mom}
    \begin{align}
        &(z - z_{cop}) \ddot{x} = (\ddot{z} + g) (x - x_{cop}),\\
        &(z - z_{cop}) \ddot{y} = (\ddot{z} + g) (y - y_{cop}).
    \end{align}
\end{subequations}

The second implication of the virtual constraint \eqref{eq:f_to_com} when combined with \eqref{eq:z_force_balance} is:
\begin{eqnarray}
\label{eq:com-cop-z}
    \ddot{z} = \omega^2 (z - z_{cop}) -g.
\end{eqnarray}
Substituting \eqref{eq:com-cop-z} back into \eqref{eq:cop-com-dynamics-w/o-ang-mom} yields:
\begin{align}
\label{eq:linear_horizontal}
    \begin{cases}
        \ddot{x} = \omega^2 (x - x_{cop}),\\
        \ddot{y} = \omega^2 (y - y_{cop}).
    \end{cases}
\end{align}
Interestingly, the virtual constraint \eqref{eq:f_to_com} leads to identical dynamical structure in all three directions, as in \eqref{eq:com-cop-z} and \eqref{eq:linear_horizontal}.
We further merge the effects of gravity and external forces through the concept of virtual repellent point (VRP) \cite{englsberger2015three},
\begin{align}
\label{def:vrp}
    \vrp &= [x_{vrp} , y_{vrp} ,z_{vrp}]^T \coloneqq \cop + [0, 0, g/\omega^2]^T,
\end{align}
and end up in the following unified dynamics in 3D:
\begin{align}
\label{eq:linear_dynamics}
    \ddot{\com} = \omega^2 (\com - \vrp).
\end{align}
It is important to note that to have valid linear 3D dynamics \eqref{eq:linear_dynamics}, it is \emph{necessary} to enforce the virtual constraint \eqref{eq:f_to_com} through feedback in the whole-body controller. The main parameter defining the virtual constraint \eqref{eq:f_to_com} is the frequency, $\omega$. In the rest of this paper, we will see that different walking and running patterns can be generated by changing $\omega$.

Similar to the 2D case, now we can split~\eqref{eq:linear_dynamics} into two first-order equations:
\begin{subequations}
\label{eq:com-dcm-dynamics}
\begin{align}
    \dot{\dcm} &= \omega(\dcm - \vrp),\label{eq:div_conv_parts_dcm}\\
    \dot{\com} &= \omega(\dcm - \com),\label{eq:div_conv_parts_com}
\end{align}
\end{subequations}
where $\dcm=[\xi_x , \xi_y , \xi_z]^T$ is the 3D-DCM \cite{englsberger2015three}. The DCM has unstable dynamics~\eqref{eq:div_conv_parts_dcm} and is pushed away by the VRP, while~\eqref{eq:div_conv_parts_com} is stable and the CoM converges to the DCM. To have a stable motion, it is enough to bound the DCM which can be done either by modulating the VRP, taking a step, or combining both. Since we are interested in proposing an approach that applies to biped robots with different ankle actuation and foot geometry (point foot, active ankle, and passive ankle), we rely only on taking steps at the desired location and time to stabilize the gait. The linear dynamics in~\eqref{eq:com-dcm-dynamics} enables us to construct an optimization problem based on \cite{khadiv2020walking} that adapts both step locations and timings for 3D walking and running through a convex optimization problem with convergence and viability guarantees.

\begin{remark}
The virtual constraint \eqref{eq:f_to_com} enforces zero angular momentum around the CoM. It is a common practice when designing trajectories using the centroidal momentum dynamics to simply minimize the angular momentum around the CoM, as angular momentum is a function of the robot's whole-body trajectories \cite{ponton2016convex,dai2016planning}. However, as explained clearly in \cite{wieber2006holonomy}, this simplifying assumption should not be interpreted as if the angular momentum is not needed for locomotion. More advanced trajectory optimization approaches use a kinematic optimizer and try to track the angular momentum and the CoM trajectories from the centroidal trajectory optimizer and alternate between them until reaching a consensus in terms of linear and angular momenta \cite{herzog2016structured,budhiraja2019dynamics,ponton2021efficient}. Interestingly, we also can adapt the virtual constraint \eqref{eq:f_to_com} by adding a time-dependent term from a full-body kinematic optimizer that accounts for the angular momentum trajectory. Hence, we believe that our approach based on virtual constraints can be used within the kino-dynamic framework \cite{herzog2016structured} likewise. Note that, as mentioned earlier, we exclude multi-contact scenarios from our analyses.
\end{remark}

\subsection{Walking and running dynamics}\label{subsec:walking-running-dynamics}
Solving~\eqref{eq:div_conv_parts_dcm} as an initial value problem in the stance phase yields:
\begin{equation}\label{dcm_stance}
    \dcm_t= e^{\omega t}(\dcm_0 -\vrpo)+\vrpo \quad ,  \quad  0 \leq t \leq T_s,	
\end{equation}
in which $\vrpo$ is the fixed VRP position in the current step and $\dcm_t$ is the DCM at time $t$.
Plugging~\eqref{dcm_stance} into~\eqref{eq:div_conv_parts_com} specifies the CoM trajectory, $\com_t$, during the stance phase, 
\begin{align}\label{com_stance}
	\com_t =& \frac{1}{2} ( \dcm_t + \vrpo + 
	e^{-\omega t} (2\com_0 {-} \dcm_0 {-} \vrpo) )
	\;, \; 0\leq t \leq T_s.
\end{align}
%
The stance phase is proceeded by a non-zero flight phase in running.   
During the flight phase, the robot's CoM follows a parabolic trajectory starting from $\com_{T_s}$ and at the initial velocity of $\dot \com_{T_s}$,
\begin{align}\label{com_flight}
    \com_t &= \frac{(t {-} T_s)^ 2}{2}  \g+ (t {-} T_s) \dot \com_{T_s}  + \com_{T_s}  \; ,  \;  T_s \leq t \leq T,
\end{align} 
where $T \coloneqq T_s + T_f$ denotes the end time of the current step.
In the absence of contact forces, the CoP cannot be defined. Still, it is facilely determined from the DCM definition per~\eqref{eq:div_conv_parts_com} that the DCM travels through a free-fall motion as well,
\begin{align}\label{dcm_flight}
    \dcm_t &= \frac{(t {-} T_s)^ 2}{2}  \g+ (t {-} T_s) (\dot \com_{T_s}  + \frac{1}{\omega}\g) + \dcm_{T_s}  \; ,  \;  T_s \leq t \leq T.	
\end{align} 

We can consider different walking and running patterns depending on the initial condition of the CoM states and the desired frequency for motions. In the rest of this paper, we only discuss running and LIPM walking, but other types of walking are imaginable, as discussed in Appendix \ref{app:modes}.
\section{Nominal Gait Generation}\label{section:nominal}
Assuming a desired average velocity, $\mathbf{v} = \begin{bmatrix} v_x, v_y,  v_z \end{bmatrix}^T$, we design a symmetric and periodic stable nominal gait characterized by the horizontal  displacement in each step, together with the stance and flight duration.
Note that this periodic gait encodes the desired behaviour that the feedback controller tries to converge to and can always be changed by the user. However, it is also possible to give the robot any arbitrary type of gait that is not periodic by specifying the desired length, width, and time of the flight and stance phases.
For simplicity, moving on flat horizontal surfaces with $v_z=0$ is considered; still, the scheme can be extended to traversing on stepping stones or uneven scenarios.
We use superscript \textit{nom} to denote nominal value for each variable. 


A periodic gait requires identical relative positions of the CoM, VRP, and DCM at the beginning of all steps,
\begin{align}\label{eq:periodic}
    \vert \dcm_{T}^{nom}-\com_{T}^{nom} \vert &= \vert \dcm_{0}^{nom}-\com_{0}^{nom} \vert,\nonumber\\
    \vert \dcm_{T}^{nom}-\vrpT^{nom} \vert &= \vert \dcm_{0}^{nom}-\vrpo^{nom} \vert.
\end{align} 
The absolute operator can be removed in \textit{x} and \textit{z} directions, but the signs must be flipped in \textit{y} coordinate. 
Additionally, moving on a flat surface implies $z_{T}^{nom}=z_{0}^{nom}$.

The nominal flight duration, $T_f^{nom}$, which brings back both the CoM and the DCM to their initial altitude, $z_{T}^{nom}=z_{0}^{nom}$ and $\xi_{z,T}^{nom}=\xi_{z,0}^{nom}$, is given by:
\begin{align}\label{T_f}
	T_f^{nom} = \frac{2 \omega(\Gamma^{nom} - 1)}{g(\Gamma^{nom} + 1)} (z_{0}^{nom} - z_{vrp,0}^{nom}),
\end{align}
where $\Gamma\coloneqq e^{\omega T_s}$.
Walking with $T_f^{nom}=0$ happens when $\omega = \omega_0 = \sqrt{g/(z_0^{nom}-z_{cop}^{nom})}$, which simplifies the general dynamics per~\eqref{eq:linear_dynamics} to LIPM. 

The nominal change between two consecutive foot step locations, denoted by $\curStep$ and $\nextStep$, on flat terrains is:
\begin{align}\label{eq:delta_vrp_nom}
    \delu^{nom} \coloneqq \nextStep^{nom} {-} \curStep^{nom} = [v_x T^{nom} , v_y T^{nom} {-} (-1)^n l_p]^T,
\end{align}
where $l_p$ is the pelvis width, $n = 1$ if the right foot is stance, and $n = 2$ otherwise. The \textit{x} and \textit{y} components of $\delu$ show step length and width, respectively.

By introducing the \textit{3D-DCM offset} as the eventual offset between the DCM and VRP at the end of a step (the 2D version has been proposed by the authors in \cite{khadiv2016step}), $\boldsymbol b \coloneqq \dcm_T - \vrpT$, and assuming the nominal step size $\delu^{nom}$, the nominal DCM offset is simply:
\begin{align}\label{b_x}
b_x^{nom} =& \frac{\Delta u_{x}^{nom}-{\dot{x}_{T_s}^{nom}}T_f^{nom}}{\Gamma^{nom}  - 1},\nonumber\\
b_y^{nom} =& \frac{(-1)^n l_p}{\Gamma^{nom}+1}+\frac{\Delta u_{y}^{nom}}{\Gamma^{nom}-1} +\nonumber \\&
\frac{(-1)^n (-\dot{y}_{T_s,r}^{nom}\Gamma^{nom}-\dot{y}_{T_s,l}^{nom})T_f}{{\Gamma^{nom}}^2-1},\nonumber\\
b_z^{nom} =&\frac{2}{\Gamma^{nom}+1}(z_0^{nom}-z_{vrp,0}^{nom}),
\end{align}
where the horizontal components of the CoM velocity at $T_s$ are conveniently described in terms of the desired velocity,
\begin{subequations}
\label{eq:dot_com_ts_nom}
\begin{align}
    \dot{x}_{T_s}^{nom} &= 
    \frac{v_x T^{nom}}{T_f^{nom} {+} 2(\Gamma^{nom}{-}1)/(\omega(\Gamma^{nom}{+}1))}
    \label{eq:dot_x_ts_nom},\\
    \dot{y}_{T_s}^{nom} &= 
    \frac{v_y T^{nom} - (-1)^n l_p}
    {T_f^{nom} {+} 2(\Gamma^{nom}{+}1)/(\omega(\Gamma^{nom}{-}1))}
    \label{eq:dot_y_ts_nom}.
\end{align}
\end{subequations}
The lateral speed depends on which foot is on the ground, as encoded by $n$ earlier, and we indicate it using subscript \textit{r} or \textit{l},  $\dot{y}_{T_s,r}$ when $n=1$ or $\dot{y}_{T_s,l}$ when $n=2$.

Interestingly, while we only imposed periodicity condition through~\eqref{eq:periodic}, the nominal CoM trajectory is symmetric as well. 
In a symmetric stance phase, the stance foot settles at an identical distance from the CoM at the beginning and end of stance, $\vert \com_{T_s}^{nom} - \vrpo^{nom} \vert = \vert \com_{0}^{nom} - \vrpo^{nom} \vert$.
The flight phase is inherently symmetric, conditioned on landing the swing foot when $z_{T}^{nom}=z_{T_s}^{nom}$. It is easy to see from~\eqref{com_stance} and~\eqref{eq:periodic} that both conditions hold (all proofs are in Appendix~\ref{proof_nominal}).

To summarize the procedure of finding nominal gait values, we assume that the user specifies a desired walking/running velocity. We use \eqref{T_f} to derive the nominal flight phase which is zero for LIPM walking. The user also needs to specify three gait hyperparameters to fully define a desired gait, i.e., frequency $\omega$, nominal stance time $T_s$ and the initial CoM height $(z_{0}^{nom} - z_{vrp,0}^{nom})$. Then, the nominal change of the foot location per step is calculated from \eqref{eq:delta_vrp_nom}. 
Finally, using \eqref{b_x}, we compute the nominal DCM offset. The main goal of our control framework is to achieve the nominal DCM offset defined in \eqref{b_x}. By trying to approach the desired DCM offset at the end of each step, we ensure that it is possible to have the desired gait in the next step in the ideal situation. More importantly, we make sure that the unstable part of the dynamics, i.e. DCM, remains bounded.

\begin{remark}
Our main result in~\cite{khadiv2020walking} shows that the viability kernel for LIPM walking can be characterized in terms of the DCM offset. This conclusion is easily extended to the more generalized dynamics of~\eqref{eq:linear_dynamics} by substituting $\omega_0$ with $\omega$ in all derivations.
For a given walking or running dynamics specified by the frequency, $\omega$, we argue that ensuring the DCM offset is inside the viability region at every single step, $\boldsymbol b < \boldsymbol b^{max}$, is sufficient for generating a stable gait: 
I) if the DCM offset is larger than $\boldsymbol b^{max}$, all possible choices of step timing and location lead to divergence,
II) otherwise, at least one combination of step timing and location keeps the DCM from diverging.
It can be confirmed from~\eqref{b_x} that a shorter stance phase and greater step size push the gait towards instability margin and make viability preservation more laborious.
\end{remark}




\section{feedback control}\label{sec:feedback-control}
Given the nominal values of the gait in the previous section, here we present a simple QP that tries to realize the nominal gait utmostly while guaranteeing the viability of the gait. This QP adapts the nominal values, e.g., step location as well as stance and flight phases duration, such that the gait remains viable in the presence of disturbances. Note that we assign the virtual constraint in \eqref{eq:f_to_com} as the main task to our whole body controller, which is equal to having no angular momentum around the CoM during motion.

The most critical constraint in our problem is the dynamics. Here, we are interested in specifying the dynamics as a function of the next step location and duration such that we can adapt position and timing based on the state measurements online.
We can express~\eqref{dcm_stance} and~\eqref{dcm_flight} in terms of the current CoM and VRP, stance and flight duration, the DCM offset, and the new VRP location:
\begin{align}\label{final_val}
	\vrpT =& \frac{1}{2} T_f^2 \g
	+ T_f(\vproj + \frac{1}{\omega} \g) + \dcmproj -\boldsymbol b.
\end{align}
In~\eqref{final_val}, $\vproj$ and $\dcmproj$ denote our belief at $t$ about the CoM velocity and DCM position at take-off time, $T_s$.
If the dynamics are studied during flight, $\vproj$ is calculated from measurement; otherwise, it is approximated by its nominal value:
\begin{subequations}\label{eq:vproj}
\begin{align}[left ={\vproj = \empheqlbrace}]
	& \dot{\com}_{T_s}^{nom} \quad &\text{if} \;t \in [0, T_s), \\
	& \dot{\com}_{t} - (t-T_s) \g \quad &\text{if} \; t \in [T_s, T).
\end{align}
\end{subequations}
The DCM position at $T_s$ is approximated by evolving current measurements back or forth in time:
\begin{subequations}\label{eq:dcmproj}
\begin{align}[left ={\dcmproj = \empheqlbrace}]
	& \Gamma e^{-\omega t} (\dcm_t - \vrpt) + \vrpt ,\quad \text{if} \; t \in [0, T_s], \\
	& \com_t - \frac{(t-T_s)^2}{2}\g + (\dot{\com}_{t} - (t-T_s) \g) \times \\
	&(\frac{1}{\omega} + T_s - t), \nonumber\; \quad \quad \quad \quad \quad \; \; \text{if} \; t \in [T_s, T].
\end{align}
\end{subequations}
In~\eqref{eq:dcmproj}, $\vrpt$ stands for the measured VRP at time $t$ which may deviate from $\vrpo$ due to external disturbances.

The other constraint is the minimum time constraint in the stance phase. For walking, this constraint limits the maximum allowable acceleration of the swing foot. In fact,  this minimum time is required to bring the swing foot from the current state to its final state because of the swing foot dynamics. For running, however, this constraint makes sure that the CoM height starts increasing during the stance phase.  Expressing this constraint  in terms of our optimization variable $\Gamma$ obtains:
\begin{align}\label{T_min}    
    T_s \geq T^{min} \implies \Gamma \geq e^{\omega T^{min}},
\end{align}
in which $T^{min}$ indicates the minimum time required for the swing foot to take a step. While for walking we use a simple fixed value as in \cite{khadiv2020walking}, for running we use the following time-dependent lower-bound to make sure that the CoM height is increasing (see Appendix \ref{app:friction} for proof) 
\begin{align}\label{T_min_run}    
    \Gamma^{min} = e^{\omega t} \sqrt{\frac{2\max\{z_t,\xi_{z,t}\} - \xi_{z,t} - z_{vrp,t}}{\xi_{z,t} - z_{vrp,t}}}.
\end{align}
%

The next constraint is the foot reachability constraint which ensures that the robot leg does not go to the kinematic singularity or the CoM does not excessively approach the ground.
The distance between two step locations, $\delu$, is either travelled through the flight phase or proceeded by stretching the leg to be landed next.
In order to prevent self collision and over-stretching of the legs, 
the second portion must lie within a feasible range: 
\begin{align}\label{cons_kin_vrp_bound}
    \delu^{min} \leq 
    \delu -  T_f \vproj^{x,y}
    \leq \delu^{max}.
\end{align}
Equation~\eqref{cons_kin_vrp_bound} limits the step length and width while allowing for bigger steps in running than in walking.
Additionally, it justifies the intuition that by increasing the stance duration while running, the robot can fly more and thus, take larger steps. 

The next kinematic constraint concerns the CoM altitude in the stance phase,
\begin{align}\label{cons_kin_z_bound}
    z^{min}\leq z_t \leq z^{max} \quad , \quad 0 \leq t \leq T_s.
\end{align}
%
The CoM goes higher as the stance time increases; hence, the maximum $\Gamma$ for not exceeding $z^{max}$ is:
\begin{align}\label{eq:T_smax}
    \Gamma \leq \Gamma^{max} = \frac{e^{\omega t}}{\xi_{z,t} - z_{vrp,t}}
    (z^{max}-z_{vrp,t} +\nonumber\\
    \sqrt{(z^{max}-\xi_{z,t})^2 +   2(z^{max}-z_t)(\xi_{z,t}-z_{vrp,t})})
\end{align}
Also, to ensure a valid height at the beginning of the next step, $T_f$ must be in $[T_{f}^{min}, T_{f}^{max}]$, where
\begin{align}\label{eq:Tfmax}
    T_{f}^{min} &{=} \max \{ t,(\sqrt{2 g (\Tilde{\xi}_{z,t} {-} \Tilde{v}_{z,t}/\omega {-} z^{max}){+} \Tilde{v}_{z,t}^2} + \Tilde{v}_{z,t})/g\},\nonumber\\
    T_{f}^{max} &{=} (\sqrt{2 g (\Tilde{\xi}_{z,t} {-} \Tilde{v}_{z,t}/\omega {-} z^{min}) {+} \Tilde{v}_{z,t}^2} + \Tilde{v}_{z,t})/g.
\end{align}
Finally, we refer the readers to Appendices~\ref{proof_constraints} and \ref{app:friction} for proofs and some considerations on the frictional constraints that are not the main concern for walking and running.


The primary equality constraint of our problem is \eqref{final_val} which is nonlinear with respect to $T_f$. To be able to adapt $T_f$ in the feedback controller while maintaining the convexity of our optimization problem, 
we only consider equations in $x,y$ directions, while the desired behaviour in the $z$ direction is achieved by steering $T_f$ to its nominal value. Now, we can construct a QP that employs the current measurements of the system and, if necessary, sacrifices the realization of the nominal motion to keep the viability of the gait. Using the current and next step locations $\curStep$ and $\nextStep$, we have:


\begin{subequations}
\label{eq:DCM_MPC}
\begin{align}
\underset{\nextStep, \Gamma, T_f, \boldsymbol b^{x,y}}{\text{minimize}} \quad 
& \alpha_1 \Vert \nextStep - \curStep - 
\delu^{nom} \Vert^2  +  \nonumber\\
&\alpha_2 |\Gamma-\Gamma^{nom}|^2 + \alpha_3 |T_f-T_f^{nom}|^2 +\nonumber\\
&\alpha_4  \Vert \boldsymbol b^{x,y} - \begin{bmatrix} b_x^{nom}\\b_y^{nom}\end{bmatrix} \Vert^2\\ \nonumber \\
\text{s.t.} \quad &\nextStep = T_f\vproj^{x,y} + \dcmproj^{x,y} -\boldsymbol b^{x,y}.\label{eq:MPC_constraint_dynamics}\\
&\nextStep - \curStep \geq \delu^{min} + T_f\vproj^{x,y}  \nonumber\\
&\nextStep - \curStep \leq  \delu^{max}+ T_f\vproj^{x,y}
\label{eq:MPC_constraint_location}\\
&\Gamma^{min}\leq \Gamma \leq \Gamma^{max}\label{eq:MPC_constraint_stance_time}\\
& T_{f}^{min} \leq T_f \leq  T_{f}^{max} \label{eq:MPC_constraint_flight_time}\\
&\begin{bmatrix} b_{x}^{min}\\b_{y,out}^{max} \end{bmatrix}\leq \boldsymbol b^{x,y} \leq \begin{bmatrix} b_{x}^{max}\\b_{y,in}^{max} \end{bmatrix} \label{eq:MPC_constraint_viability}
\end{align}
\end{subequations}
Due to self-collision, the bounds in the lateral direction are not symmetric and we define $b_{y,in}$ and $b_{y,out}$ as in~\cite{khadiv2020walking}.

The first three terms in the cost function try to bring the next step location and timing to their nominal values. The last term encourages the DCM offset towards the nominal DCM offset and is given a larger weight compared to the other terms such that the optimizer adapts the current gait parameters to ensure a \emph{viable} next step. The system dynamics are encoded in~\eqref{eq:MPC_constraint_dynamics}, and~\eqref{eq:MPC_constraint_location} ensures that the kinematic reachability limitations of the next step are respected. \eqref{eq:MPC_constraint_stance_time} and \eqref{eq:MPC_constraint_flight_time} are constraints on the stance and swing time. Finally, \eqref{eq:MPC_constraint_viability} guarantees the boundedness of the DCM offset, which maintains the viability of the gait. This small QP can be solved in a fraction of ms on a standard laptop using an off-the-shelf QP solver \cite{goldfarb1983numerically}.



\section{RESULTS}
In this section, we present the results of applying our proposed controller for walking and running of the biped robot Bolt \cite{daneshmand2021variable} in Pybullet \cite{coumans2016pybullet}. Bolt is a fully open-source biped robot with passive ankles and 3 active degrees of freedom per leg.

All simulations are carried out on a laptop with 2.8 GHz Core i7 processor and 16GB RAM. We use the whole-body controller in \cite{grimminger2020open}. The main tracking tasks we have are CoM control in $z$ direction and the base roll and pitch angles to keep the base as upright as possible. These three tasks are enough to minimize the deviations from the virtual constraint \eqref{eq:f_to_com}. As Bolt's ankles are passive, we do not control the CoM or DCM in horizontal directions and rely only on \eqref{eq:DCM_MPC} to adapt step location and timing based on measurements to stabilize the motion. We carry out different walking and running simulation experiments with different velocities where the command velocity changes during walking and running. We also applied external disturbances and performed walking on random uneven terrains to show the robustness of our control framework. All simulations are included in the accompanying video\footnote{ \url{https://www.youtube.com/watch?v=Chz3CGDNkRQ}}.

\begin{figure*}
	\centering
	\includegraphics[height=0.17\linewidth, trim={0cm 0cm 0cm 0cm}, clip]{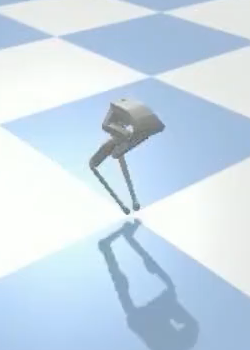}%
	\includegraphics[height=0.17\linewidth, trim={0cm 0cm 0cm 0cm}, clip]{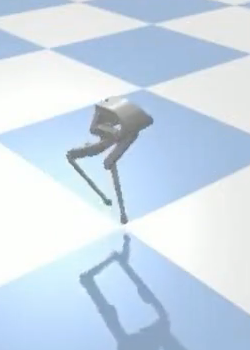}%
	\includegraphics[height=0.17\linewidth, trim={0cm 0cm 0cm 0cm}, clip]{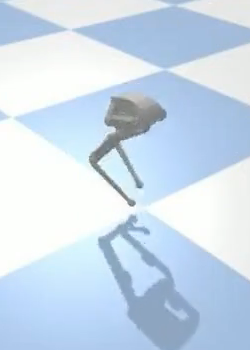}%
	\includegraphics[height=0.17\linewidth, trim={0cm 0cm 0cm 0cm}, clip]{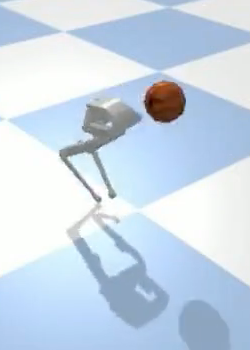}%
	\includegraphics[height=0.17\linewidth, trim={0cm 0cm 0cm 0cm}, clip]{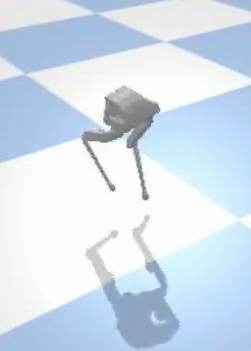}%
	\includegraphics[height=0.17\linewidth, trim={0cm 0cm 0cm 0cm}, clip]{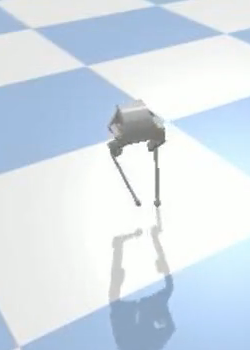}%
	\includegraphics[height=0.17\linewidth, trim={0cm 0cm 0cm 0cm}, clip]{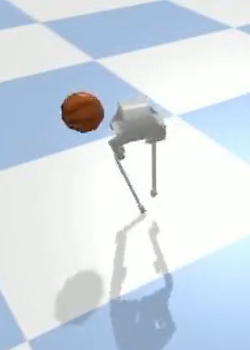}%
	\includegraphics[height=0.17\linewidth, trim={0cm 0cm 0cm 0cm}, clip]{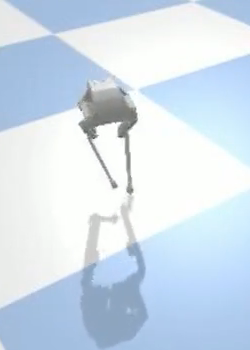}%
	\vspace{-3mm}
	\caption[]{\small Snapshots of the simulation scenario (left to right). The robot starts from running and then switches to walking in place. The robot is pushed during running by random disturbances (a 0.1 Kg ball is thrown towards the robot base) at $t=0.95$ sec and during walking at $t=2.5$ sec and recover from these pushes by adapting both step location and timing.}
	\label{fig:snapshots}
	\vspace{-5mm}
\end{figure*}

\begin{figure}
\centering
\includegraphics[width=\linewidth]{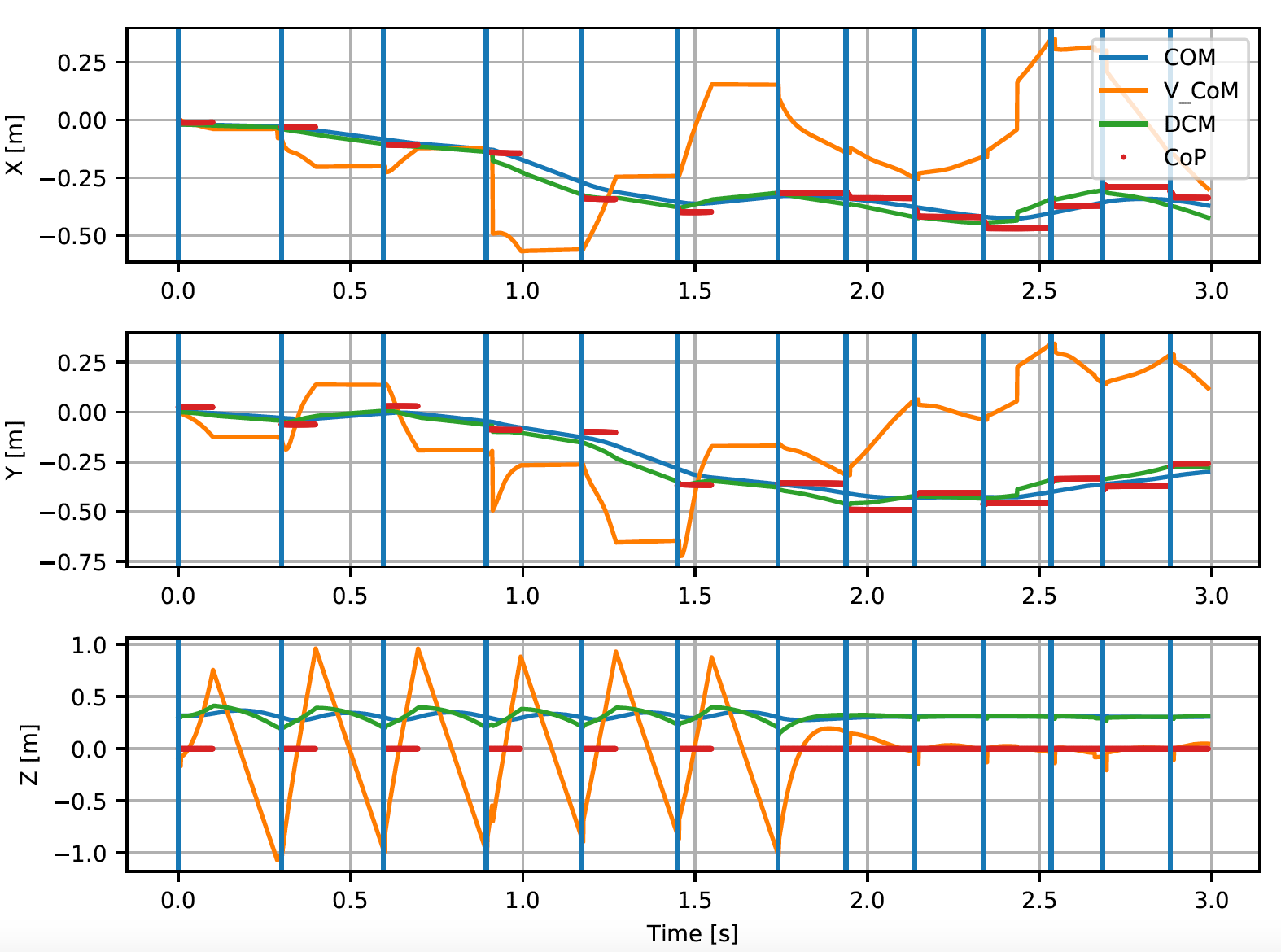}
\vspace{-8mm}
\caption{Simulation in Pybullet. Bolt starts with running at the beginning of this scenario and then switches to walking. The robot is pushed during motion by a random disturbance (a ball is thrown towards the robot base in the simulation) and adapts both step location and timing to recover.}
\label{walking-running}
\vspace{-5mm}
\end{figure}

\begin{figure}
\centering
\includegraphics[width=\linewidth]{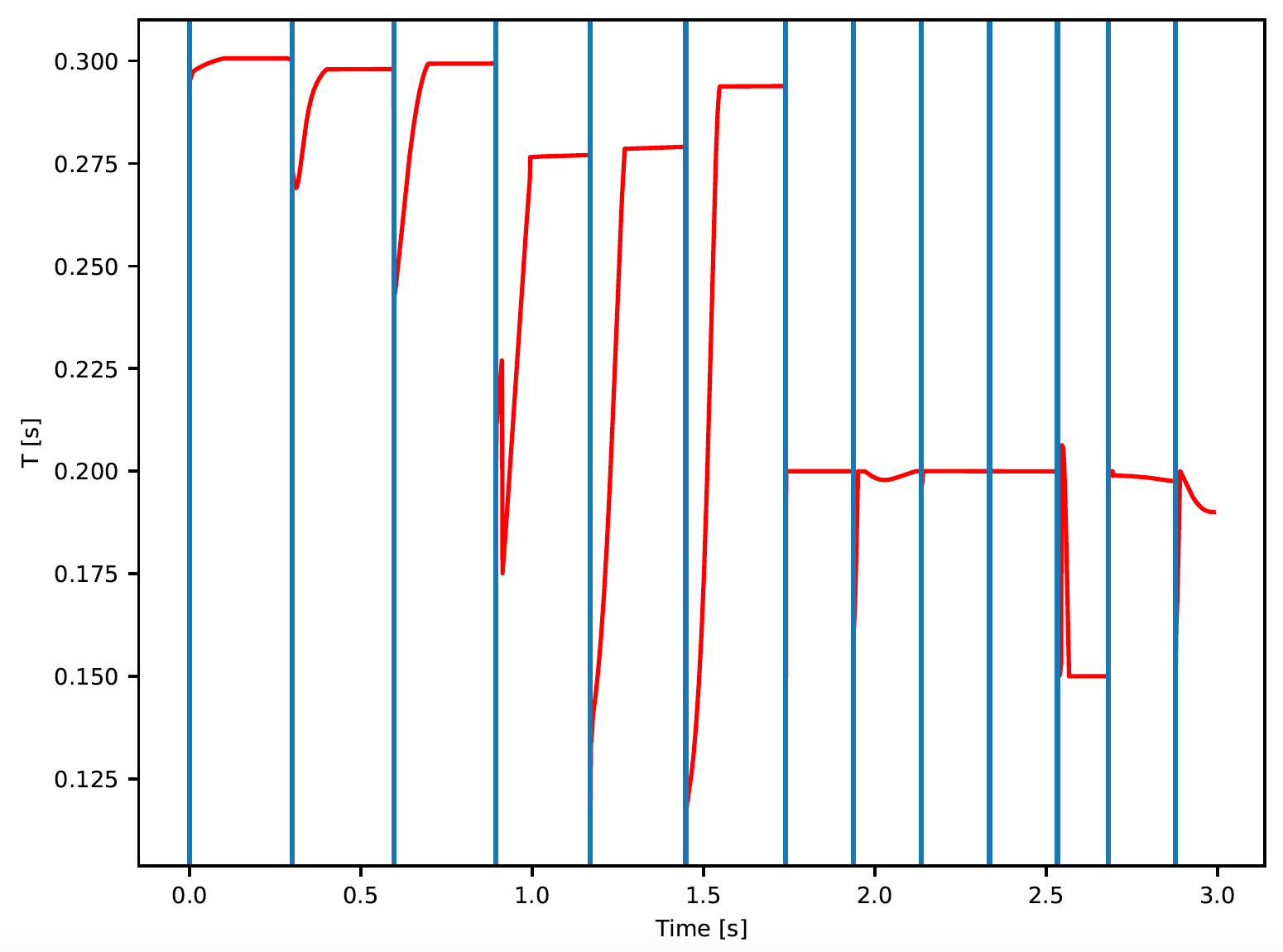}
\vspace{-8mm}
\caption{Adaptation of step timing in the presence of external disturbances.}
\label{time-adaptation}
\vspace{-5mm}
\end{figure}

To show adaptation of step location and time, here we present the result of a simulation scenario where the robot performs a combination of walking and running (see Fig. \ref{fig:snapshots}), while it is pushed by an external force. Figure \ref{walking-running} presents the evolution of the CoM, DCM, as well as CoP during this scenario. The robot starts from running and then switches to walking in place. The robot is pushed during running by random disturbances (a 0.1 Kg ball is thrown towards the robot base, see the accompanying video) at $t=0.95$ sec and during walking at $t=2.5$ sec and recover from these pushes by adapting both step location and timing. To better illustrate adaptation of the step time, we also show the step time $T$ of the controller \eqref{eq:DCM_MPC} for this scenario in Fig. \ref{time-adaptation}. As it can be observed in this figure, step timing is adapted to complement step location adjustment for push recovery, immediately after the robot is pushed.

\section{conclusions and future work}
In this paper, we presented a unified framework to design and control walking and running gaits for bipedal robots. The central concept to our framework is the notion of virtual constraint in the centroidal space. We have shown that we can generate different walking and running motions and control them through a single walking controller. Our results showed that, a simple QP to adapt next step location and timing can enable our biped robot Bolt to robustly walk and run over uncertain surfaces and in the presence of external disturbances.

In our future work, we will try to extend our framework for walking and running on structured 3D environments such as staircases. Furthermore, we are interested in combining our low-dimensional predictive controller with a whole body model predictive control (MPC) framework where the goal is to have a consensus in terms of step location and timing. Finally, realizing running motions on the real Bolt is our next goal.




\bibliography{Master}

\begin{thebibliography}{10}
\providecommand{\url}[1]{#1}
\csname url@rmstyle\endcsname
\providecommand{\newblock}{\relax}
\providecommand{\bibinfo}[2]{#2}
\providecommand\BIBentrySTDinterwordspacing{\spaceskip=0pt\relax}
\providecommand\BIBentryALTinterwordstretchfactor{4}
\providecommand\BIBentryALTinterwordspacing{\spaceskip=\fontdimen2\font plus
\BIBentryALTinterwordstretchfactor\fontdimen3\font minus
  \fontdimen4\font\relax}
\providecommand\BIBforeignlanguage[2]{{%
\expandafter\ifx\csname l@#1\endcsname\relax
\typeout{** WARNING: IEEEtran.bst: No hyphenation pattern has been}%
\typeout{** loaded for the language `#1'. Using the pattern for}%
\typeout{** the default language instead.}%
\else
\language=\csname l@#1\endcsname
\fi
#2}}

\bibitem{kajita20013d}
S.~Kajita, F.~Kanehiro, K.~Kaneko, K.~Yokoi, and H.~Hirukawa, ``The 3d linear
  inverted pendulum mode: A simple modeling for a biped walking pattern
  generation,'' in \emph{Proceedings 2001 IEEE/RSJ International Conference on
  Intelligent Robots and Systems. Expanding the Societal Role of Robotics in
  the the Next Millennium (Cat. No. 01CH37180)}, vol.~1.\hskip 1em plus 0.5em
  minus 0.4em\relax IEEE, 2001, pp. 239--246.

\bibitem{blickhan1989spring}
R.~Blickhan, ``The spring-mass model for running and hopping,'' \emph{Journal
  of biomechanics}, vol.~22, no. 11-12, pp. 1217--1227, 1989.

\bibitem{takenaka2009real}
T.~Takenaka, T.~Matsumoto, and T.~Yoshiike, ``Real time motion generation and
  control for biped robot-1 st report: Walking gait pattern generation,'' in
  \emph{2009 IEEE/RSJ International Conference on Intelligent Robots and
  Systems}.\hskip 1em plus 0.5em minus 0.4em\relax IEEE, 2009, pp. 1084--1091.

\bibitem{hubicki2018walking}
C.~Hubicki, A.~Abate, P.~Clary, S.~Rezazadeh, M.~Jones, A.~Peekema, J.~Van~Why,
  R.~Domres, A.~Wu, W.~Martin, \emph{et~al.}, ``Walking and running with
  passive compliance: Lessons from engineering: A live demonstration of the
  atrias biped,'' \emph{IEEE Robotics \& Automation Magazine}, vol.~25, no.~3,
  pp. 23--39, 2018.

\bibitem{daneshmand2021variable}
E.~Daneshmand, M.~Khadiv, F.~Grimminger, and L.~Righetti, ``Variable horizon
  mpc with swing foot dynamics for bipedal walking control,'' \emph{IEEE
  Robotics and Automation Letters}, vol.~6, no.~2, pp. 2349--2356, 2021.

\bibitem{chen2020optimal}
Y.-M. Chen and M.~Posa, ``Optimal reduced-order modeling of bipedal
  locomotion,'' in \emph{2020 IEEE International Conference on Robotics and
  Automation (ICRA)}.\hskip 1em plus 0.5em minus 0.4em\relax IEEE, 2020, pp.
  8753--8760.

\bibitem{herzog2016structured}
A.~Herzog, S.~Schaal, and L.~Righetti, ``Structured contact force optimization
  for kino-dynamic motion generation,'' in \emph{2016 IEEE/RSJ International
  Conference on Intelligent Robots and Systems (IROS)}.\hskip 1em plus 0.5em
  minus 0.4em\relax IEEE, 2016, pp. 2703--2710.

\bibitem{dai2016planning}
H.~Dai and R.~Tedrake, ``Planning robust walking motion on uneven terrain via
  convex optimization,'' in \emph{2016 IEEE-RAS 16th International Conference
  on Humanoid Robots (Humanoids)}.\hskip 1em plus 0.5em minus 0.4em\relax IEEE,
  2016, pp. 579--586.

\bibitem{carpentier2018multicontact}
J.~Carpentier and N.~Mansard, ``Multicontact locomotion of legged robots,''
  \emph{IEEE Transactions on Robotics}, vol.~34, no.~6, pp. 1441--1460, 2018.

\bibitem{wieber2006holonomy}
P.-B. Wieber, ``Holonomy and nonholonomy in the dynamics of articulated
  motion,'' in \emph{Fast motions in biomechanics and robotics}.\hskip 1em plus
  0.5em minus 0.4em\relax Springer, 2006, pp. 411--425.

\bibitem{ponton2021efficient}
B.~Ponton, M.~Khadiv, A.~Meduri, and L.~Righetti, ``Efficient multicontact
  pattern generation with sequential convex approximations of the centroidal
  dynamics,'' \emph{IEEE Transactions on Robotics}, 2021.

\bibitem{sleiman2021unified}
J.-P. Sleiman, F.~Farshidian, M.~V. Minniti, and M.~Hutter, ``A unified mpc
  framework for whole-body dynamic locomotion and manipulation,'' \emph{IEEE
  Robotics and Automation Letters}, vol.~6, no.~3, pp. 4688--4695, 2021.

\bibitem{shah2021rapid}
P.~Shah, A.~Meduri, W.~Merkt, M.~Khadiv, I.~Havoutis, and L.~Righetti, ``Rapid
  convex optimization of centroidal dynamics using block coordinate descent,''
  in \emph{2021 IEEE/RSJ International Conference on Intelligent Robots and
  Systems (IROS)}, 2021.

\bibitem{westervelt2018feedback}
E.~R. Westervelt, J.~W. Grizzle, C.~Chevallereau, J.~H. Choi, and B.~Morris,
  \emph{Feedback control of dynamic bipedal robot locomotion}.\hskip 1em plus
  0.5em minus 0.4em\relax CRC press, 2018.

\bibitem{reher2016realizing}
J.~Reher, E.~A. Cousineau, A.~Hereid, C.~M. Hubicki, and A.~D. Ames,
  ``Realizing dynamic and efficient bipedal locomotion on the humanoid robot
  durus,'' in \emph{2016 IEEE International Conference on Robotics and
  Automation (ICRA)}.\hskip 1em plus 0.5em minus 0.4em\relax IEEE, 2016, pp.
  1794--1801.

\bibitem{nguyen2020dynamic}
Q.~Nguyen, X.~Da, J.~Grizzle, and K.~Sreenath, ``Dynamic walking on stepping
  stones with gait library and control barrier functions,'' \emph{Algorithmic
  Foundations of Robotics XII. Springer}, pp. 384--399, 2020.

\bibitem{reher2021control}
J.~Reher and A.~D. Ames, ``Control lyapunov functions for compliant hybrid zero
  dynamic walking,'' \emph{arXiv preprint arXiv:2107.04241}, 2021.

\bibitem{sreenath2013embedding}
K.~Sreenath, H.-W. Park, I.~Poulakakis, and J.~W. Grizzle, ``Embedding active
  force control within the compliant hybrid zero dynamics to achieve stable,
  fast running on mabel,'' \emph{The International Journal of Robotics
  Research}, vol.~32, no.~3, pp. 324--345, 2013.

\bibitem{ameshybrid}
A.~D. Ames and I.~Poulakakis, ``Hybrid zero dynamics control of legged
  robots,'' 2018.

\bibitem{guo2021fast}
Y.~Guo, M.~Zhang, H.~Dong, and M.~Zhao, ``Fast online planning for bipedal
  locomotion via centroidal model predictive gait synthesis,'' \emph{IEEE
  Robotics and Automation Letters}, 2021.

\bibitem{orin2013centroidal}
D.~E. Orin, A.~Goswami, and S.-H. Lee, ``Centroidal dynamics of a humanoid
  robot,'' \emph{Autonomous robots}, vol.~35, no.~2, pp. 161--176, 2013.

\bibitem{wieber2016modeling}
P.-B. Wieber, R.~Tedrake, and S.~Kuindersma, ``Modeling and control of legged
  robots,'' in \emph{Springer Handbook of Robotics}.\hskip 1em plus 0.5em minus
  0.4em\relax Springer, 2016, pp. 1203--1234.

\bibitem{englsberger2015three}
J.~Englsberger, C.~Ott, and A.~Albu-Sch{\"a}ffer, ``Three-dimensional bipedal
  walking control based on divergent component of motion,'' \emph{IEEE
  Transactions on Robotics}, vol.~31, no.~2, pp. 355--368, 2015.

\bibitem{khadiv2020walking}
M.~Khadiv, A.~Herzog, S.~A. Moosavian, and L.~Righetti, ``Walking control based
  on step timing adaptation,'' \emph{IEEE Transactions on Robotics}, 2020.

\bibitem{ponton2016convex}
B.~Ponton, A.~Herzog, S.~Schaal, and L.~Righetti, ``A convex model of momentum
  dynamics for multi-contact motion generation,'' \emph{arXiv preprint
  arXiv:1607.08644}, 2016.

\bibitem{budhiraja2019dynamics}
R.~Budhiraja, J.~Carpentier, and N.~Mansard, ``Dynamics consensus between
  centroidal and whole-body models for locomotion of legged robots,'' in
  \emph{2019 International Conference on Robotics and Automation (ICRA)}.\hskip
  1em plus 0.5em minus 0.4em\relax IEEE, 2019, pp. 6727--6733.

\bibitem{khadiv2016step}
M.~Khadiv, A.~Herzog, S.~A.~A. Moosavian, and L.~Righetti, ``Step timing
  adjustment: A step toward generating robust gaits,'' in \emph{Humanoid Robots
  (Humanoids), 2016 IEEE-RAS 16th International Conference on}.\hskip 1em plus
  0.5em minus 0.4em\relax IEEE, 2016, pp. 35--42.

\bibitem{goldfarb1983numerically}
D.~Goldfarb and A.~Idnani, ``A numerically stable dual method for solving
  strictly convex quadratic programs,'' \emph{Mathematical programming},
  vol.~27, no.~1, pp. 1--33, 1983.

\bibitem{coumans2016pybullet}
E.~Coumans and Y.~Bai, ``Pybullet, a python module for physics simulation for
  games, robotics and machine learning,'' \emph{GitHub repository}, 2016.

\bibitem{grimminger2020open}
F.~Grimminger, A.~Meduri, M.~Khadiv, J.~Viereck, M.~W{\"u}thrich, M.~Naveau,
  V.~Berenz, S.~Heim, F.~Widmaier, T.~Flayols, \emph{et~al.}, ``An open
  torque-controlled modular robot architecture for legged locomotion
  research,'' \emph{IEEE Robotics and Automation Letters}, vol.~5, no.~2, pp.
  3650--3657, 2020.

\bibitem{walkingdef}
D.~L. Gebo, ``Functional vertebrate morphology.'' \emph{American Journal of
  Physical Anthropology}, vol.~72, no.~1, pp. 131--132, 1987.

\bibitem{froud}
R.~Alexander, ``Optimization and gaits in the locomotion of vertebrates,''
  \emph{Physiological reviews}, vol.~69, no.~4, p. 1199—1227, 1989.

\bibitem{Fr_human}
R.~Alexander and McN., ``Mechanics and scaling of terrestrial locomotion,''
  1977, p. 93–110.

\bibitem{Fr_birds}
S.~Gatesy and A.~Biewener, ``Bipedal locomotion: effects of speed, size and
  limb posture in birds and humans,'' \emph{Journal of Zoology}, vol. 224, pp.
  127 -- 147, 2009.

\end{thebibliography}
\bibliographystyle{IEEEtran}

\appendix
\subsection{Modes of motion}\label{app:modes}
Based on our formulation, we can categorize walking and running in four groups and call them \textit{modes of motion} according to the CoM altitude pattern.
The first three modes, \textit{LIPM walking}, \textit{walking}, and \textit{sneaking}, do not include a flight phase, $T_f=0$, while the last one is \textit{running} and involves non-zero flight phases.
We argue that each mode is associated with a frequency range and transiting between different modes is done by changing $\omega$ in consecutive steps (note that $\omega$ is kept constant during a single step).


Walking is classically defined as a gait in which at least one leg is in contact with the ground at all times~\cite{walkingdef}.
A simplifying assumption is to maintain a constant CoM height which is the principal assumption of the well-known LIPM \cite{kajita20013d}.
Using the natural frequency, $\omega=\omega_0=\sqrt{g/(z_0-z_{cop})}$ in~\eqref{def:vrp} yields $z_{vrp,0} = z_0$ that boils down~\eqref{com_stance} to $z_t=z_0$.
The general dynamics per~\eqref{eq:linear_dynamics} is reduced to LIPM in this case, and we call the corresponding motion \textit{LIPM walking}.

If $\omega <\omega_0$, it follows from~\eqref{def:vrp} that the CoM is initially located below the VRP, $z_0 < z_{vrp,0}$, rises from $t{=}0$ to $t{=}T_s/2$, and then falls until it reaches the initial height, $z_{T_s} = z_0$, forming a sequence of vaults centered at contact points.
The described gait is the most compatible with human walking with stretched legs, hence we call it simply \emph{walking}. 
In addition to the well-studied LIPM walking and the general walking gait, we introduce another subdivision of walking that we refer to as \textit{sneaking}.
Repeating the same procedure for $\omega > \omega_0$ shows that $z_0 - z_{vrp,0} > 0$; so, the CoM goes down at first and then rises.
This motion can be useful as an intermediate gait between walking and running. 
It is important to note that for both \emph{walking} and \emph{sneaking}, a double support phase is essential to bring the CoM states at the end of current stance phase to the beginning of the next stance phase.

Any motion with a non-zero flight duration in which both feet are lifted up, is referred to as running.
The stance dynamics resembles a spring (in vertical direction) continuing to go down after landing each foot for $T_s/2$, getting the closest to the ground straightly above the contact point before being pushed up and away from the stance foot for the second half of the period. 
At the end of the stance phase, the foot is lifted up and the CoM goes through a ballistic trajectory.
To identify the concordant frequency range, note that the flight duration per~\eqref{T_f} must be positive, or equivalently $\dcm_{z,0} - z_{vrp,0} > 0$, which is only plausible when $\omega>\omega_0$. 

Terrestrial locomotors decide between walking and running based on the motion velocity, simultaneously transiting from walking to running as the pace increases.
Biologist have investigated various structural and metabolic triggers to explain the speed threshold between walking and running; though, none could concretely justify the speed-up and down switching ensemble.
A dimensionless measure in continuum mechanics referred to as \textit{Froude number (Fr)} is propitious in quantifying the transition speed~\cite{froud}.
Defined as $Fr \coloneqq (v_x^2 + v_y^2)/(g \, d)$, where $d$ denotes the CoM distance from the contact point, Froude number builds upon the absolute velocity to leg length ratio.
While walking at any Froude number smaller than $1$ is mathematically plausible, observations suggest a switching point of $Fr=0.5$ for many biped species such as humans~\cite{Fr_human} and birds~\cite{Fr_birds}.
However, in this paper we do not take these considerations into account and implement walking and running for different walking velocities from zero to their maximum range. We believe that measures like energy efficiency and robustness need to be taken into account to find an optimal gait for different situations.

\subsection{Friction cone constraints} \label{app:friction} 
To avoid stance foot slippage, the contact forces must always lie inside the friction cone,
\begin{align}\label{frictioncone}
	\sqrt{f_x^2+f_y^2} \leq \mu_s  f_z ,
\end{align}
where $\mu_s$ is the static friction coefficient between the surface and the stance foot.
Using~\eqref{eq:f_to_com} to relate the external forces to the CoM location during stance,~\eqref{frictioncone} is rewritten as:
\begin{align}\label{app_friction_eq1}
    \sqrt{(x_t - x_{vrp})^2 + (y_t - y_{vrp})^2} \leq
    \mu_s  z_t  \, , \, 0\leq t \leq T_s.
\end{align}
The beginning and end of a step are the critical times, when the sides of~\eqref{app_friction_eq1} are the closest to each other.
Assuming that the stance foot does not slip initially, ~\eqref{frictioncone} establishes another upper bound on $\Gamma_s$.

\subsection{Proofs for Section~\ref{section:nominal}}\label{proof_nominal}
For fixed $\dcm_0$ and $\vrpo$, applying the periodicity condition~\eqref{eq:periodic} restricts possible CoM stance trajectories in~\ref{com_stance} to: 
\begin{align}\label{com_stance_nominal}
	\com_t^{nom} =& 0.5 (e^{\omega t} + \Gamma e^{-\omega t}\boldsymbol\alpha)(\dcm_0- \vrpo) + 
	\vrpo,
\end{align}
where $\boldsymbol\alpha=[-1,1,1]^T$.
The CoM trajectory~\eqref{com_stance_nominal} is symmetric w.r.t the VRP.

\subsubsection{Proof of Eq.~\eqref{b_x}}\label{app:b_x} 
By the DCM offset definition,
\begin{align*}
b_x = & 
\xi_{x,T}-x_{vrp,T} = (\xi_{x,T} {-} x_{vrp,0}) - \Delta u_x \quad \textit{(by~\eqref{def:vrp})}\\	
= &  T_f \dot{x}_{T_s} + \Gamma (\xi_{x,0}-x_{vrp,0})  - \Delta u_x	\quad\textit{(by~\eqref{dcm_stance},\eqref{dcm_flight})}\\
= & T_f \dot{x}_{T_s} + \Gamma b_x   - \Delta u_x		\quad\textit{(by gait periodicity~\eqref{eq:periodic})}.
\end{align*}
The last line holds since the DCM offset is the same in all steps and obtains the formula of~\eqref{b_x}.

The proof in the lateral direction is similar, but the effect of pelvis length must be taken into account. Repeating the same procedure as for $b_x$,
\begin{align*}
    v_y T + l_p - \dot{y}_{T_s,r}\, T_f = \Gamma\, b_{y,r} - b_{y,l},\nonumber\\
    v_y T - l_p - \dot{y}_{T_s,l} \,T_f = \Gamma\, b_{y,l} - b_{y,r},
\end{align*}
which together obtain $b_y$.

Finally, Eq.~\eqref{com_stance_nominal} at $t=0$ is used for calculating $b_z$:
\begin{align*}
    &z_0 = 0.5(\Gamma + 1) (\xi_{z,0}-z_{vrp,0})+z_{vrp,0} \nonumber \\
    \implies
    &\xi_{z,0}-z_{vrp,0} = 0.5(\Gamma + 1) (z_0-z_{vrp,0}).
\end{align*}
The last line equals $b_z$ because of periodicity~\eqref{eq:periodic}

\subsubsection{Proof of Eq.~\eqref{eq:dot_com_ts_nom}}
We prove~\eqref{eq:dot_y_ts_nom} (\eqref{eq:dot_x_ts_nom} is similar).
First, use~\eqref{def:vrp},~\eqref{eq:periodic} and~\eqref{com_stance_nominal}, to find the nominal step width:
\begin{align*}
    \Delta u_y^{nom} &= (y_{vrp,T}{+}y_{vrp,0})-2 \, y_{vrp,0}\nonumber \\
    &= (y_{T}{+}y_{0})-2 \, y_{vrp,0}\nonumber \\
    &= T_f \dot{y}_{T_s} + (\Gamma + 1) (\xi_{y,0} - y_{vrp,0})\nonumber\\
    &= T_f \dot{y}_{T_s} + \frac{2(\Gamma + 1)}{\omega (\Gamma-1)}\dot{y}_{T_s}^{nom}.
\end{align*}
The last line is obtained from taking the derivative of ~\eqref{com_stance_nominal} and yields~\eqref{eq:dot_y_ts_nom} together with~\eqref{eq:delta_vrp_nom}.

\subsection{Proofs for Section~\ref{sec:feedback-control}}\label{proof_constraints}
Most equations in this section involve approximating the CoM or DCM position at $T_s$ from measurements at $t$. A general trick is to shift~\eqref{dcm_stance} and~\eqref{com_stance} in time such that current measurements are used instead of initial values,
\begin{align}\label{eq:com_Ts}
    \com_{T_s} &= \frac{1}{2}(\Gamma e^{-\omega t} (\dcm_{t} {-} \vrpt) \nonumber\\ &+\Gamma^{-1} e^{\omega t} (2 \, \com_t {-}\dcm_{t} {-}\vrpt)) 
    +\vrpt.
\end{align}

\subsubsection{Proof of~\eqref{final_val} to~\eqref{eq:dcmproj}}
Eq.~\eqref{final_val} is derived by replacing $\dot{\com}_{T_s}$ and $\dcm_{T_s}$ with their approximation at $t$. For proving~\eqref{eq:vproj} and~\eqref{eq:dcmproj}, compare~\eqref{dcm_stance}-\eqref{dcm_flight} at $t{=}t$ and $t{=}T_s$.

\subsubsection{Proof of~\eqref{T_min_run}}\label{proof:T_min_run}
If $\dot{z}_t\geq0$, the upward-moving requirement is already satisfied and the stance phase can end immediately, $\Gamma_{min}=e^{\omega t}$.
Otherwise, $\xi_{z,t} < z_t$ and setting $\dot{z}_{T_s} \geq 0$ in the time derivative of~\eqref{eq:com_Ts} obtains:
\begin{align*}  
    \Gamma_{min} = e^{\omega t} \sqrt{(2 z_t - \xi_{z,t} - z_{vrp,t})/(\xi_{z,t} - z_{vrp,t})}.
\end{align*}
The two cases based on $\dot{z}_t$ can be summarised as in~\eqref{T_min_run}.

%

%
\subsubsection{Proof of Eq.~\eqref{eq:T_smax}}
Solving $z_{T_s}=z_{max}$ in~\eqref{eq:com_Ts} obtains $\Gamma_{max}$ per~\eqref{eq:T_smax}, but the validity of the answer must be checked, i.e. \textit{i)} $\Gamma_{max} \geq e^{\omega t}$, and \textit{ii)} the term under the square root should be non-negative.
By triangle inequality,
\begin{align*}
    \Gamma_{max} \geq e^{\omega t} \frac{\vert z_{max} - \xi_{z,t} \vert + z_{max} - z_{vrp,t}}{\xi_{z,t} - z_{vrp,t}}\geq e^{\omega t}.
\end{align*}
Since $\Ddot{z}_t\geq 0$, $z_t \geq z_{vrp,t}$ and $\xi_{z,t} \geq z_{vrp,t}$, so \textit{ii} holds.

\subsubsection{Proof of Eq.~\eqref{eq:Tfmax}}
In~\eqref{com_flight}, approximate $\dot{\com}_{T_s}$ and $\com_{T_s}$ by $\vproj$ and $\dcmproj-\vproj/\omega$. Then solve $z_{T}=z_{max}$ to get $T_{f,min}$ (similarly for $T_{f,max}$).

\end{document}